\documentclass[letterpaper, 10 pt, conference]{ieeeconf}  
\usepackage{graphicx}
\usepackage{booktabs}
\usepackage{amsmath}
\usepackage{amssymb}
\usepackage{siunitx}
\usepackage{multirow}

\IEEEoverridecommandlockouts                              

\title{\LARGE \bf
ODeform: Learning Continuous 4D Motion for Shape Deformation \\ with Neural ODEs
}

\author{Yordanka Velikova$^{1,2}$, Mahdi Saleh$^{1}$, Liming Kuang$^{1,2}$, Benjamin Busam$^{1,2}$ 
\thanks{$^{1}$ Technical University of Munich}%
\thanks{$^{2}$ Munich Center for Machine Learning (MCML)
        }%
}

\begin{document}

\maketitle
\thispagestyle{empty}
\pagestyle{empty}

\begin{abstract}
Modeling continuous object deformation is important for many computer vision and robotics tasks, such as manipulation and simulation. Existing approaches rely on learning-based methods or physics simulators to model shape deformations. However, these approaches either use discrete time steps or are too computationally intensive for real-time applications. We present ODeform, a novel extension of Neural Ordinary Differential Equations to continuous 4D dynamics of deformable objects in 3D space.
Our method transforms 3D point clouds and physical conditions (like material properties) into a unified latent space. By solving the resulting ordinary differential equations over time, we model deformations as continuous flows within this learned embedding, eliminating the need for discrete time steps while maintaining computational efficiency.
We evaluate our approach on unseen physical parameter configurations, showing improved motion prediction accuracy over baseline methods. Our experiments further demonstrate a successful transfer to real 3D captured objects with novel shapes, along with effective interpolation and extrapolation of the learned dynamics.
Our code and data will be made publicly available.\footnote{https://github.com/danivelikova/odeform}
\end{abstract}




\section{Introduction}


Understanding and predicting object deformation is essential for robotic perception and manipulation. Interactions are continuous processes, influenced by both object's intrinsic material properties and external environmental forces. Real-world actions can include a ball bouncing off a surface, a bottle being grasped, or an object deforming under tactile interactions such as poking. Modeling these dynamic phenomena requires representations that capture both 3D shape and temporal evolution. While significant advancements have been achieved in representing static 3D shapes, predicting dynamic object changes remains challenging. 

\label{sec:intro}
\begin{figure}
    \centering
    \includegraphics[width=\linewidth]{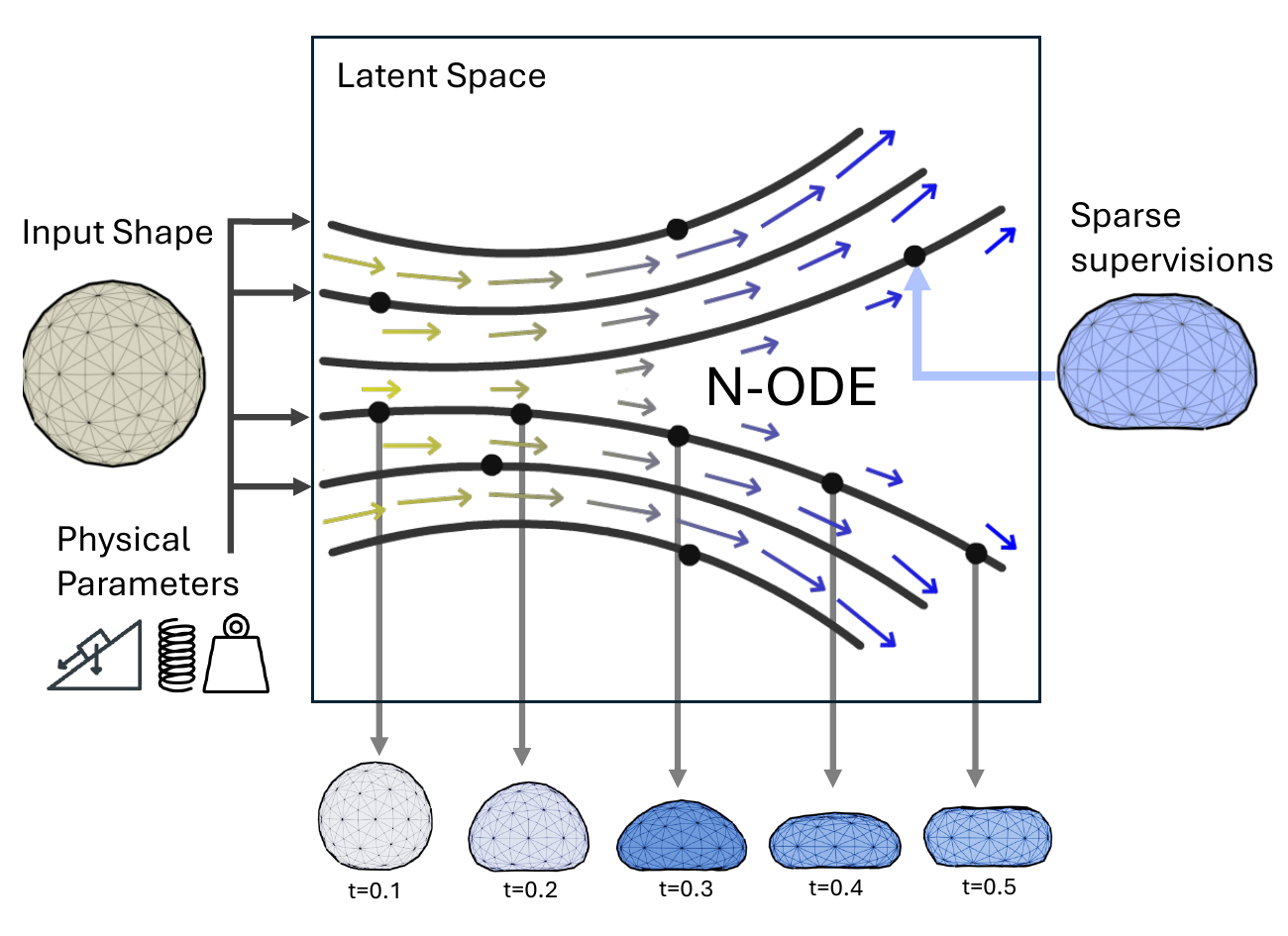}
    \caption{Overview of our continuous deformation modeling framework. Given an input 3D shape and physical parameters, our method learns a latent space representation where neural ODEs (nODEs) model the continuous evolution of object deformation over time. ODeform can generate intermediate deformation states at arbitrary timesteps while only requiring sparse supervisory signals. The learned latent space enables smooth interpolation between states while maintaining physical consistency.}
    \label{fig:teaser}
\end{figure}

Classical physics-based simulators solve partial differential equations using finite element methods (FEM)~\cite{Pan2015SubspaceDS, Barbi2005RealTimeSI, pharr2023physically}, providing accurate results grounded in physical principles. However, their computational demands and sensitivity to configuration make them impractical for real-time robotic applications, particularly with geometrically complex or dense objects. 
Learning-based alternatives, including point-based methods~\cite{liu2019meteornet,vu2022rfnet,liu2019flownet3d} and autoregressive approaches~\cite{jin2022deformation}, predict deformations directly from observations but operate on discrete time steps, producing physically implausible intermediate states and struggling to generalize to unseen geometries or material properties.

Neural Ordinary Differential Equations (nODEs)~\cite{chen2018neural} offer a promising direction for modeling continuous-time dynamics. Rather than advancing state at fixed intervals, nODEs define system evolution through a continuous learned vector field, allowing integration at arbitrary time points. This enables true interpolation between observed states and extrapolation into future dynamics, capabilities directly relevant to predictive robotic control. Crucially, because nODEs assume deterministic motion from a fixed initial state, they are naturally suited to physical deformation tasks where each configuration has a unique outcome.

While Neural ODEs have successfully modeled continuous dynamics in low-dimensional spaces, extending this paradigm to 3D deformations present significant theoretical challenges. 
By design, nODEs inherently assume deterministic motion, as their formulation relies on a single, fixed vector field for each initial state. This is particularly important for physical systems with deterministic dynamics, where each initial state has exactly one outcome, making nODEs well-suited for object deformation tasks. 
Based on these foundations, we propose ODeform, a novel framework that leverages nODEs to model continuous 4D dynamics of deformable objects.
As shown in Figure~\ref{fig:teaser}, our approach encodes point clouds and physical parameters into a latent representation evolved through time via a learned ODE function. 
Our work addresses limitations in current ODE formulations by introducing a novel decomposition of the dynamics field while enabling efficient integration.
In summary, our key contributions are:
\begin{itemize}
    \item A novel neural ODE-based architecture integrating geometric and physical priors for \textbf{continuous-time deformation modeling of 3D objects.}
    \item \textbf{Generalizable dynamics model} transferring across object geometries and unseen physical conditions while maintaining physical consistency.
    \item \textbf{Continuous dynamics representation} enabling interpolation and extrapolation across multiple 3D formats such as point clouds, meshes, gaussian splats.    
\end{itemize}

\section{Related Works}
\label{sec:related_works}
\subsection{Classical Deformation Modeling}
Physics-based simulators model deformations by solving differential equations using finite element methods (FEM)~\cite{pharr2023physically}, providing physically accurate results grounded in well-established mechanics. However, their computational demands and sensitivity to expert configuration make them impractical for real-time robotic applications, particularly when interacting with geometrically complex or dense objects. Efforts to mitigate computational demands include kinematic simplifications and low-dimensional mappings~\cite{sharp2023data}, which help approximate complex systems more efficiently but at the cost of accuracy or generality.

Learning-based methods have emerged as faster alternatives, predicting deformations directly from observations. Point-based approaches~\cite{liu2019meteornet,vu2022rfnet} model scene flow across discrete time steps, while autoregressive models~\cite{kerr2024robot} iteratively predict future states. Specialized models like PE-GNN~\cite{saleh2024physics} encodes physical parameters into graph nodes for contact-driven deformation prediction, and 3D-PhysNet~\cite{wang20183d} learns material properties via adversarial training. Despite these advances,such approaches rely on discrete time steps, resulting in physically implausible intermediate states, and often overfit to specific observations rather than modeling the full temporal dynamics required for dynamic robotic manipulation
\subsection{Deformable Object Interaction in Robotics}
Robotic manipulation of deformable objects presents unique challenges, as objects alter shape upon contact in ways that are difficult to anticipate and control~\cite{arriola2020modeling}. Early work addressed this through subspace simulation methods that merge physics-based and learning-based techniques to approximate contact-driven deformations efficiently~\cite{romero2021learning, romero2022contact}. For hand-object interaction, hyperbolic space representations have been explored to reconstruct deforming meshes from image observations~\cite{leng2023dynamic}. Tactile-based approaches~\cite{wang2022tacto} leverage contact sensing for deformation estimates during manipulation, coupling perception with physical interaction. Model-based controllers~\cite{navarro2016automatic, ficuciello2018fem} incorporate deformation predictions into closed-loop robotic control. However, these methods typically assume known material properties, simplified geometries, or controlled lab settings, and lack the ability to generalize during deployment.  

\subsection{Continuous Dynamic Modeling}
Continuous representations address the limitations of both discrete modeling and restricted generalization. Geometry-based methods build on the as-rigid-as-possible paradigm~\cite{sorkine2007rigid}, providing strong geometric priors but limited physical realism for complex contact scenarios. Implicit representations enable resolution-independent modeling: Occupancy Flow~\cite{niemeyer2019occupancy} learns continuous 4D vector fields for particle deformation, while neural fields~\cite{yang2021geometry} allow interpolation between deformation states. While expressive, these methods lack a principled mechanism for modeling how deformation evolves continuously under varying physical conditions.

Neural Ordinary Differential Equations (nODEs)~\cite{chen2018neural, bilovs2021neural} offer such a mechanism, parameterizing system derivatives with neural networks and enabling continuous-time modeling without fixed time discretization. A key characteristic is their inherently deterministic nature, future states are fully determined by current conditions, which aligns naturally with physical deformation processes that follow deterministic rules of mechanics. CaSPR~\cite{rempe2020caspr} demonstrates the promise of this direction by canonicalizing point spaces and applying nODEs for continuous time representation of 3D shapes; however, it lacks integration of physical parameters. Other works has expanded nODE capabilities through graph neural networks for relational dynamics~\cite{poli2019graph}, physics-inspired training constraints~\cite{greydanus2019hamiltonian}, and static-dynamic component separation~\cite{auzina2024modulated}. ODE2VAE~\cite{yildiz2019ode2vae} infers continuous dynamics from 2D image sequences, EulerFlow~\cite{vedder2024eulerflow} applies nODEs to continuous space-time scene flow, and hybrid approaches~\cite{hofherr2023neural} integrate implicit learning with explicit physical parameters for deterministic motion. However, most existing approaches either operate in lower-dimensional spaces or fail to leverage the full potential of nODEs for modeling complex deformations in 3D, while generalizing across different physical configurations.

Unlike prior methods, ODeform leverages neural ODEs to model continuous 4D deformation in a learned latent space. By jointly encoding geometry and physical parameters, our approach enables interpolation, extrapolation, and generalization to unseen configurations. This continuous formulation provides a foundation for applications where temporal consistency and physical accuracy are important.




\begin{figure*}
    \centering
    \includegraphics[width=1.0\textwidth]{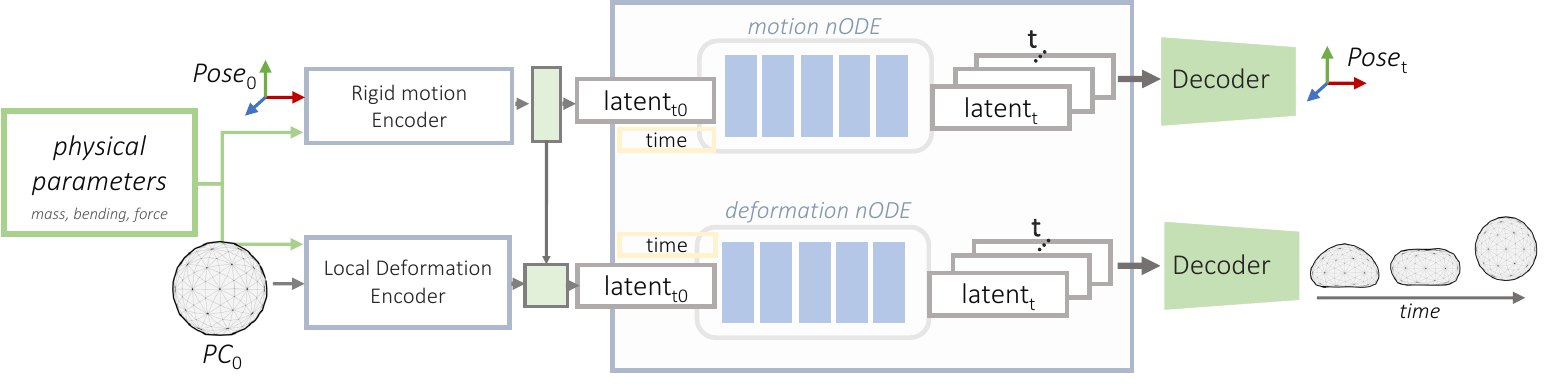}
    \caption{Overview of ODeform architecture for continuous deformation modeling. The framework consists of three main components: (1) A dual encoder that processes the initial point cloud ($PC_0$) and physical parameters (mass, elasticity, force) through separate pathways for rigid motion and local deformation, (2) Two parallel neural ODEs that evolve the latent representations continuously in time, capturing global pose changes and local shape deformations separately, and (3) Specialized decoders that reconstruct the final state, producing both rigid transformation ($Pose_t$) and local shape changes at arbitrary timestamps. The neural ODEs enable smooth interpolation between discrete observations while preserving physical consistency through the learned dynamics.}
    \label{fig:pipeline}
\end{figure*}

\section{Methodology}
\label{sec:method}
Our architecture comprises three key components: (1) a dual encoder that processes geometry and physical parameters, (2) parallel neural ODEs for global and local dynamics, (3) specialized decoders for motion reconstruction. The complete pipeline is illustrated in Fig.~\ref{fig:pipeline}.
At its core, our method decomposes object dynamics into two parallel continuous flows, one for rigid body motion and another for local deformation. 
The integration of both components happens during the final reconstruction phase, preserving their distinct continuous dynamics during the learning process.

\subsection{Theoretical Foundations}
Neural ODEs~\cite{chen2018neural} are a class of continuous-time models  that describe the evolution of a hidden state \( h(t) \) using an ordinary differential equation (ODE):

\begin{equation}
\frac{d h(t)}{dt} = f_{\theta}(h(t), t), \quad \text{with} \quad h(t_0) = h_0
\label{eq:ode}
\end{equation}

where, \( f_{\theta} \) is a neural network parameterized by \( \theta \), which governs the state dynamics of over time. Unlike discrete models, nODEs enable the continuous evaluation of  \( h(t) \) at any arbitrary time step by solving the ODE numerically:
\begin{equation}
h_0, \dots, h_N = \text{ODESolve}(f_{\theta}, h_0, (t_0, \dots, t_N)).
\end{equation}

However, modeling continuous dynamics of deformable objects in 3D space presents mathematical challenges not addressed by standard nODEs, 
which model system dynamics through a single vector field $f$ as shown in Equation~\eqref{eq:ode}. 
When modeling both rigid motion $T(t)$ and local deformation $D(t)$ jointly:
\begin{equation}
\frac{d}{dt}(T(t) \circ D(t)(P_0)) \neq f(T(t) \circ D(t)(P_0), t)
\end{equation}
the inequality highlights that the derivative of the composed transformation cannot be directly modeled as a function of the transformed point cloud. The composed derivative requires chain rule application, introducing complex interactions between global and local motion that destabilize training.
We addresses these issues by modeling the continuous evolution of rigid transformations and local deformations  as separate, parallel neural ODEs.
This formulation offers several mathematical advantages: (1) it preserves SE(3) structure in the rigid component, (2) it separates motion scales to improve numerical stability, and (3) it allows specialized architectures for each component. Furthermore, this decomposition naturally aligns with the underlying physics—rigid body dynamics and material deformation properties follow different governing principles that are better captured separately.

\subsection{Continuous Deformation Modeling}
Given a temporal sequence of point clouds $\{\mathcal{P}_t\}_{t=0}^T$ representing a deforming object and physical parameters $\theta$, we aim to learn a continuous representation of its dynamics. Each point cloud $\mathcal{P}_t = \{p_i^t\}_{i=1}^N \in \mathbb{R}^{N \times 3}$ consists of $N$ points at time $t$. Our goal is to learn a mapping $\Phi: (\mathcal{P}_0, \theta, t) \rightarrow \mathcal{P}_t$ that predicts object configurations at any time $t$ given only initial state and parameters.

\subsection{Motion Decomposition}
In physical interactions, objects often undergo both rigid displacement and local deformation simultaneously. The mapping $\Phi$ captures this into a composition of both: 
\begin{equation}
   \Phi(\mathcal{P}_0, \theta, t) = T(t) \circ D(t)(\mathcal{P}_0)
\end{equation}
where $T(t) \in SE(3)$ represents global rigid transformation and $D(t): \mathbb{R}^3 \rightarrow \mathbb{R}^3$ captures local shape deformation. 
This decomposition enables natural integration of physical constraints in the rigid motion component. For example, when a soft ball bounces, $T(t)$ captures its parabolic trajectory while $D(t)$ models compression at impact. This principled decomposition guides both our architecture design and learning objectives.




\subsection{Physics and Geometry Encoding}
To implement motion decomposition, we first employ a dual encoding architecture.
We encode the initial configuration into a latent representation where the initial state and physical parameters go through two pathways:
\begin{align}
    z_g &= E_g(\mathcal{P}_0, \theta) \in \mathbb{R}^{32}, \quad
    z_l = E_l(\mathcal{P}_0, \theta) \in \mathbb{R}^{128}
\end{align}
The global encoder, $E_g$, processes the initial object pose $\mathcal{P}_0$ and physical parameters $\theta$ to produce a low-dimensional representation $z_g$ capturing the overall rigid transformation. 
In parallel, the local deformation encoder $E_l$ processes the initial point cloud $\mathcal{P}_0$ and physical parameters $\theta$ to produce a higher-dimensional latent $z_l$ representing local deformation. 
In practice, we implement both encoders using 5-layer multilayer perceptrons (MLPs) to learn dense mappings from the input space to their respective latent dimensions. While $E_g$ outputs a compact 32-dimensional representation focused on rigid motion parameters, $E_l$ projects to a larger 128-dimensional space to capture the complexity of local deformations.
Additionally we concatenate the latent of the global encoder $z_g$  with the local latent codes $z_l$ before inputing it to the neural ODE model. 




\subsection{Latent Neural ODE}

Following equation~\eqref{eq:ode}, 
we model the temporal evolution of the latent states for both global and local components using a system of two parallel neural ODEs:
\begin{align}
    \text{Global motion: } &\frac{dz_g(t)}{dt} = f_g(z_g(t), t)\\
    \text{Local deformation: } &\frac{dz_l(t)}{dt} = f_l(z_l(t), t)
\end{align}
where the initial conditions for these ODEs are provided by the respective encoders: $z_g(0) = E_g(\mathcal{P}_0, \theta)$ and $z_l(0) = E_l(\mathcal{P}_0, \theta)$.
This formulation allows us to solve for any intermediate state $\mathcal{P}(t)$ by integrating the learned dynamics. 
Training leverages the adjoint sensitivity method, which computes gradients by solving a companion ODE backward in time~\cite{Chen2018NeuralOD} and the solution to the ODEs is computed using an adaptive Runge-Kutta 4 solver. 






\subsection{Decoder and loss functions}


\textbf{State Recovery:}
We decode the latent trajectories $z_g(t)$ and $z_l(t)$ back into the original object configurations to compute the final state at any time $t$ through a decoder $D$:
\begin{equation}
\mathcal{P}(t) = D(z_g(t), z_l(t))
\end{equation}

where $D$ comprises two MLPs: $D_g$ maps $z_g(t)$ to rigid transformation parameters $T(t) \in SE(3)$, and $D_l$ transforms $z_l(t)$ to point-wise displacements $D(t)$. We use 3-layer MLPs combined with ReLU activation functions, effectively learning the inverse mappings from latent space to motion trajectories.


\noindent\textbf{Loss Functions:}
To train our model end-to-end, we use an automatically weighted loss inspired by \cite{Liebel2018AuxiliaryTI}, which dynamically adjusts the balance between learning the rigid and deformable aspects of the object's motion. The total loss is a weighted sum of a rigid transformation loss $L_{rigid}$ and a point cloud reconstruction loss $L_{points}$. The weights, \( \lambda_1 \) and \( \lambda_2 \), are optimised at each time step to account for differences in loss magnitude and variance across tasks, ensuring accurate learning of both global and local motion:
\begin{equation}
    L_\text{total} = \lambda_1 L_\text{rigid} + \lambda_2 L_\text{points}
\end{equation}


\subsection{Model Application}
During inference, our model operates in three steps. First, the dual encoders process the initial point cloud $\mathcal{P}_0$ and physical parameters $\theta$ to produce global and local latent representations $z_g(0)$ and $z_l(0)$. Second, the latent states are evolved through their respective neural ODEs to any desired timestamp $t$. Finally, the decoder networks $D_g$ and $D_l$ transform these trajectories into rigid motion $T(t)$ and local deformations $D(t)$, enabling prediction $\Phi(\mathcal{P}_0, \theta, t)$ at arbitrary temporal resolution.
\begin{figure}
    \centering
    \includegraphics[width=1.0\linewidth]{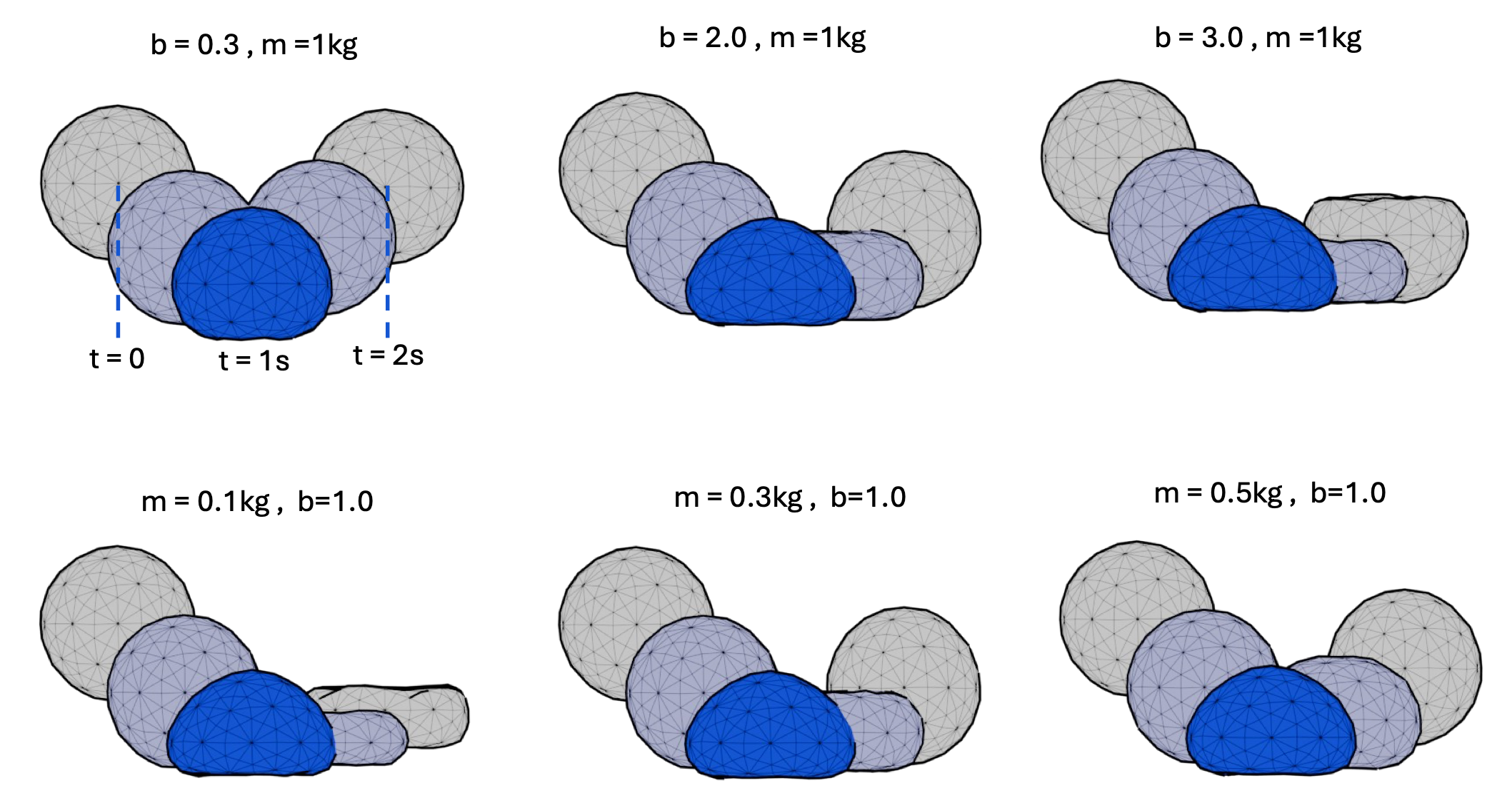}
    \caption{Effect of mass and bending on object deformation. Inference results from our model, demonstrating our method's accuracy under diverse physical parameters. Top: fixed mass (1 kg), varying elasticity (0.3–3.0). Bottom: fixed bending (1), varying mass (0.1–0.5 kg).} 
    \label{fig:fix_mass_bending}
\end{figure}

\section{Evaluation}
\label{sec:eval}
We evaluate our method through extensive experiments demonstrating its generalization capabilities across varying physical parameters and geometric configurations. Our evaluation focuses on three key aspects: (1) performance on unseen physical parameters, (2) performance on interpolation and extrapolation tasks, and (3) adaptation to different 3D representations, including gaussian splatting for reconstruction applications.

\subsection{Datasets and Simulation}
Our evaluation utilizes two complementary datasets designed to test different aspects of deformation modeling: multi-object contact interactions and mass-dependent elastic deformations. 

\begin{table*}[t]
\centering
\caption{Comparison of deformation prediction accuracy on unseen physical parameters. Results show RMSE (mm), MAE (mm), and MSE (mm²) for ODeform and baseline methods across two test conditions: Unseen Contact Force and Unseen Mass, Bending parameters. Lower values indicate better performance.}
\setlength{\tabcolsep}{6pt} 
\resizebox{0.8\textwidth}{!}{
\begin{tabular}{lSSS SSS} \toprule
\multirow{2}{*}{\textbf{Method}} & \multicolumn{3}{c}{\textbf{Unseen Contact Force}} & \multicolumn{3}{c}{\textbf{Unseen Mass, Bending}} \\ \cmidrule(lr){2-4} \cmidrule(lr){5-7}
 & {\textbf{RMSE} (mm) $\downarrow$} & {\textbf{MAE} (mm) $\downarrow$} & {\textbf{MSE} (mm$^2$) $\downarrow$} & {\textbf{RMSE} (mm) $\downarrow$} & {\textbf{MAE} (mm) $\downarrow$} & {\textbf{MSE} (mm$^2$) $\downarrow$} \\ \midrule
nODE & 6.750 & 4.956 & 82.710 & 10.344 & 6.366 & 205.000 \\
RNN & 22.670 & 16.076 & 683.000 & 13.585 & 9.035 & 260.000 \\
PEGNN~\cite{Saleh2024PhysicsEncodedGN} & 14.220 & 9.837 & 244.400 & {-} & {-} & {-} \\
\textbf{ODeform} & {\bfseries 6.657} & {\bfseries 5.140} & {\bfseries 58.430} & {\bfseries 1.279} & {\bfseries 0.948} & {\bfseries 2.000} \\ \bottomrule
\end{tabular}
}
\label{tab:comp_with_baselines_results}
\end{table*}

\textbf{Contact-Force Dataset}
We extend the Everyday Deform Dataset~\cite{Saleh2024PhysicsEncodedGN}, which contains seven distinct objects with varying material properties and mesh complexities sourced from 3D scans and CAD models. While the original dataset only provided resting and the final deformation state, we enhanced it with simulated interactions between rigid projectiles and soft objects using Blender's physics engine. 
Our enhanced dataset generates 2.1k unique deformation sequences through rigid-soft body interactions. In our setup, soft bodies are positioned at the coordinate origin with fixed material parameters, while rigid projectiles are initialized randomly near the surface with varying velocities and force vectors. Each interaction is recorded for 20 frames, capturing the complete deformation dynamics. We store contact point coordinates and force vectors alongside the soft body parameters. The resulting dataset comprises approximately 42k meshes across 7 object classes, divided into a 70/10/20 split for training, validation, and testing. 

\textbf{Mass-Elastic Dataset}
Our second dataset focuses on controlled drop experiments to analyze mass-dependent deformation dynamics. The experimental setup consists of drops from a height of 0.5m under standard gravity (9.81 m/s²). We parameterize the mass ($m$) and bending stiffness ($EI$) during the simulation. Mass affects both gravitational force and impact deformation characteristics, while bending stiffness controls the object's resistance to flexural deformation. These parameters are randomly sampled to generate diverse deformation behaviors, ranging from rigid-like responses to highly elastic deformations during impact.
Each sequence records deformation dynamics for 10 seconds. The dataset contains 500 sequences, split into 350/50/100 for training, validation, and testing. Each sequence captures the complete dynamics from the initial fall through impact and subsequent bounces. 

\subsection{Evaluation Metrics}
We evaluate our method using standard error metrics between predicted and ground truth point positions: Mean Squared Error (MSE), Root Mean Squared Error (RMSE), and Mean Absolute Error (MAE). MSE and RMSE are particularly sensitive to large deviations, while MAE provides a more uniform assessment of prediction errors, making these metrics complementary for evaluating our method's accuracy across different scales of deformation.

\subsection{Generalization to new physical parameters}
\label{exp_mass_bending}
We test on a subset of unseen physical parameter configurations to evaluate our model's ability to handle novel physical properties. We consider parameters including material properties (mass, bending) for the case of mass-elastic deformation and contact conditions (contact point, force vector) for contact-force deformation.
\begin{table}[ht]
\centering
\caption{Results on the Contact-Force test set data. Comparison of deformation prediction for each object and time point.}
\resizebox{\linewidth}{!}{
\begin{tabular}{lSSS} \toprule
    \textbf{Method} & {\textbf{RMSE} (mm) $\downarrow$} & {\textbf{MAE} (mm) $\downarrow$} & {\textbf{MSE} (mm$^2$) $\downarrow$} \\ \midrule
    \text{Bottle} & 7.322 & 5.733 & 65.000\\
    \text{Cat} & 9.298 & 7.203 & 108.000 \\
    \text{Dog} & 10.617 & 8.380 & 128.000 \\
    \text{Donut} & 4.058 & 3.000 & 18.000\\
    \text{Doritos} & 4.842 & 3.676 & 26.000  \\
    \text{Flipflop} & 5.760 & 4.428 & 39.000 \\
    \text{Pillow} & 4.705 & 3.560 & 25.000 \\ \midrule
    \textbf{Average} & {\bfseries 6.657} & {\bfseries 5.140} & {\bfseries 58.430} \\ \bottomrule
\end{tabular}}
\label{tab:everyday_objects_eval_odeform}
\end{table}



\noindent To evaluate generalization to novel physical scenarios, we first test ODeform on unseen contact point deformations across seven objects from the Contact-Force dataset (Table~\ref{tab:everyday_objects_eval_odeform}). The model achieves the best performance on the Donut object (RMSE: 4.058mm, MAE: 3.0mm, MSE: 18.0mm²), while maintaining consistent performance across other household objects like Bottles, Pillows, and Flipflop (RMSE: 4.705-7.322mm), suggesting model's practical applicability to real-world complex geometries.

We evaluate our method's generalization capabilities on two test conditions (Table~\ref{tab:comp_with_baselines_results}): unseen force vectors in the Contact-Force dataset and unseen mass-bending parameters in the Mass-Elastic dataset. We compare against three baselines: a neural ODE trained directly on point clouds without latent encoding, a recurrent neural network (RNN), and PE-GNN~\cite{Saleh2024PhysicsEncodedGN}. While PE-GNN leverages graph networks for structural relationships, it only predicts final states. The neural ODE and RNN baselines, though capable of temporal modeling, struggle with shape preservation and geometric details respectively.

\begin{figure}
\centering
    \includegraphics[width=1.0\linewidth]{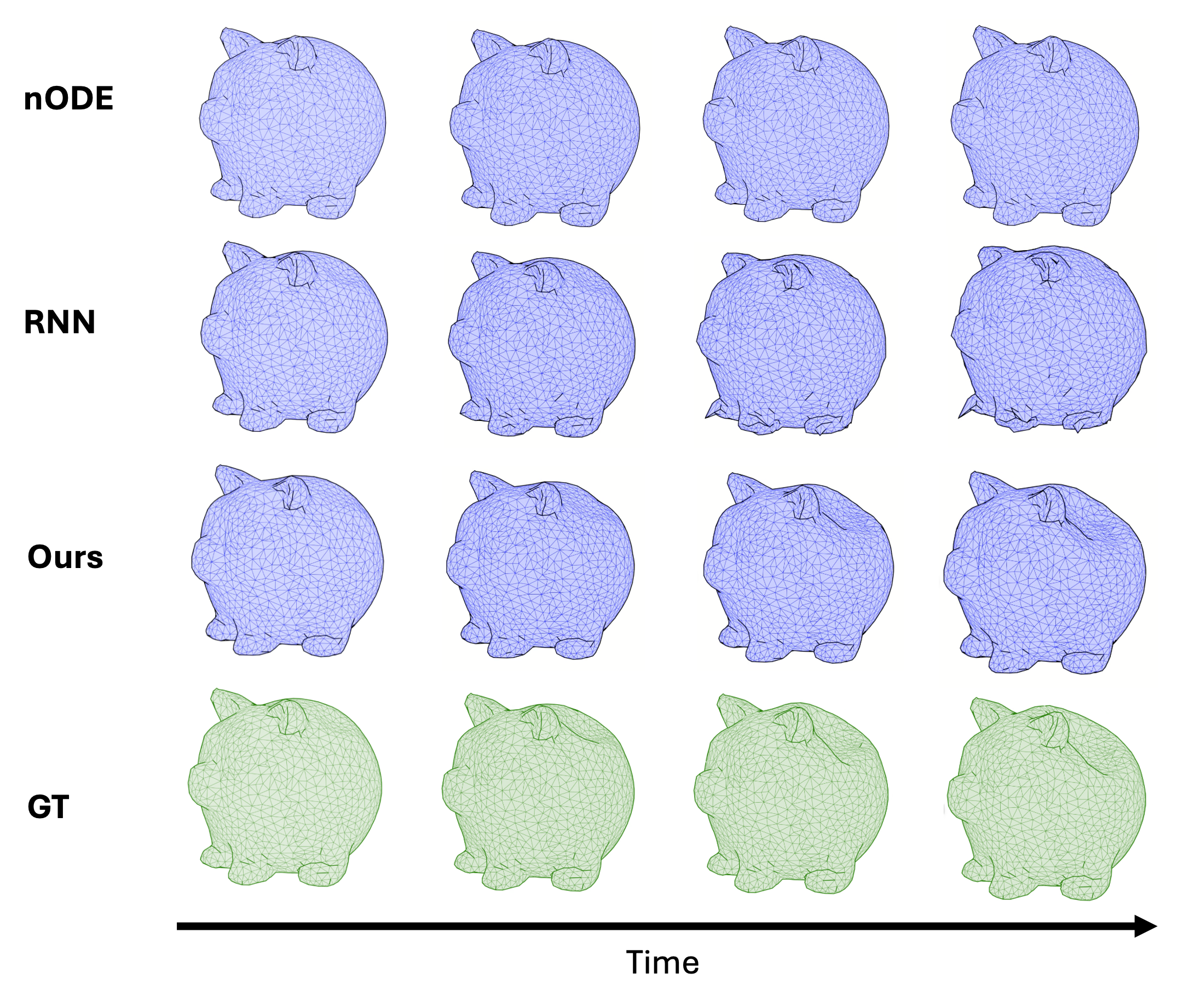}
    \caption{Comparison of deformation predictions on object from the Contact-Force dataset. Our dual nODE approach better models deformation of unseen force vector over time compared to baseline methods, achieving results closer to ground truth (GT).} 
    \label{fig:baseline-results-quant}
\end{figure}

Our approach achieves the lowest errors across all metrics, with RMSE of 6.657 mm for unseen contact forces and 1.279 mm for unseen mass-bending parameters. Figure~\ref{fig:baseline-results-quant} demonstrates the generalization of local deformation of a force vector applied with time.
While baselines exhibit over-smoothing (nODE) or shape instability (RNN), our method can model the deformation and maintain shape consistency that aligns with ground truth.

In Figure~\ref{fig:fix_mass_bending}, we present our model's performance on the Mass-Elastic dataset under varying physical parameters. With fixed mass (1 kg), we vary bending (0.3-3.0) to demonstrate material flexibility effects on fall dynamics. Similarly, with fixed bending, we vary mass (0.1-0.5 kg) to showcase mass-dependent deformation patterns. These results validate our method's ability to handle diverse physical parameters while maintaining physically plausible deformations, supporting its applicability in scenarios requiring parameter handling.

\subsection{Performance on Unseen Geometry}
\begin{figure}
    \centering
    \includegraphics[width=1.0\linewidth]{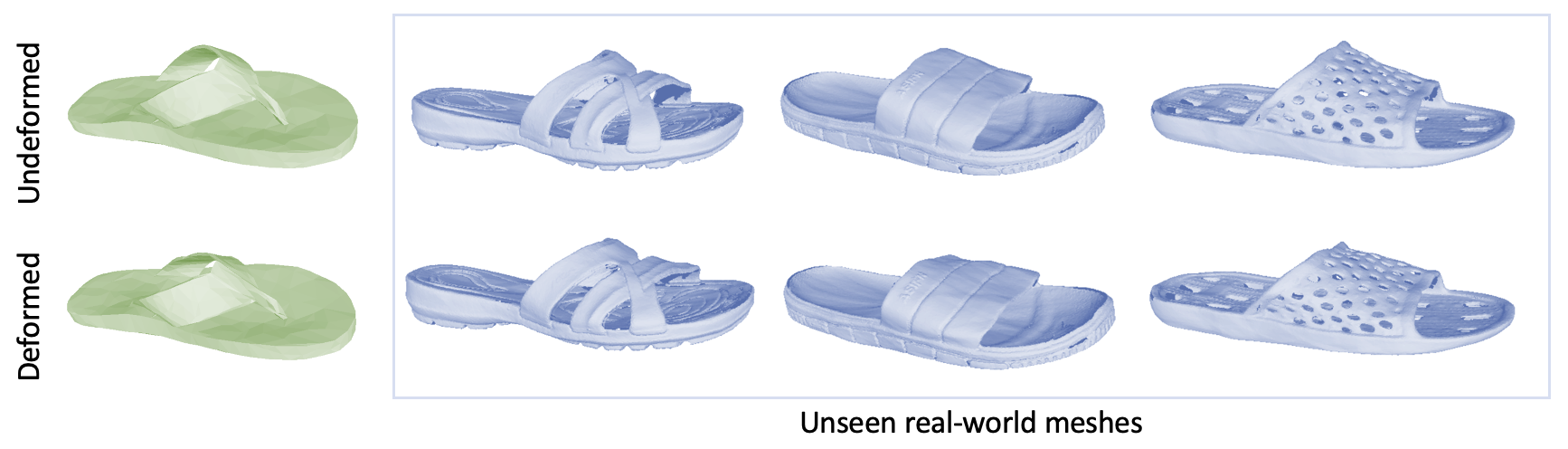}
    \caption{Deformation to real-world captured meshes. The comparison shows undeformed (top) and deformed (bottom) states of unseen real-world objects processed by our model, demonstrating successful transfer of learned deformation dynamics to new instances.} 
    \label{fig:housecat}
\end{figure}

We evaluate our model's ability to transfer learned deformation dynamics to real-world captures of novel instances from the same category. Using the HouseCAT6D~\cite{jung2024housecat6d} dataset, which contains real-world captured objects across different classes, we showcase the results on the shoe category, as it is contained in both datasets. Figure \ref{fig:housecat} demonstrates how our model, trained on synthetic shapes from the Contact-Force dataset,  transfers the learned motion to  reconstructed meshes from the HouseCAT6D dataset. 
The comparison shows undeformed (top) and deformed (bottom) states of
unseen real-world objects processed by our model, demonstrating successful transfer of learned deformation dynamics to new instances, compared to ground truth deformation from the synthetic example. 
The results illustrating our method's ability to translate structurally consistent deformations on novel instances.

\subsection{Gaussian Splatting Integration}
\begin{figure}
    \centering
    \includegraphics[width=1.0\linewidth]{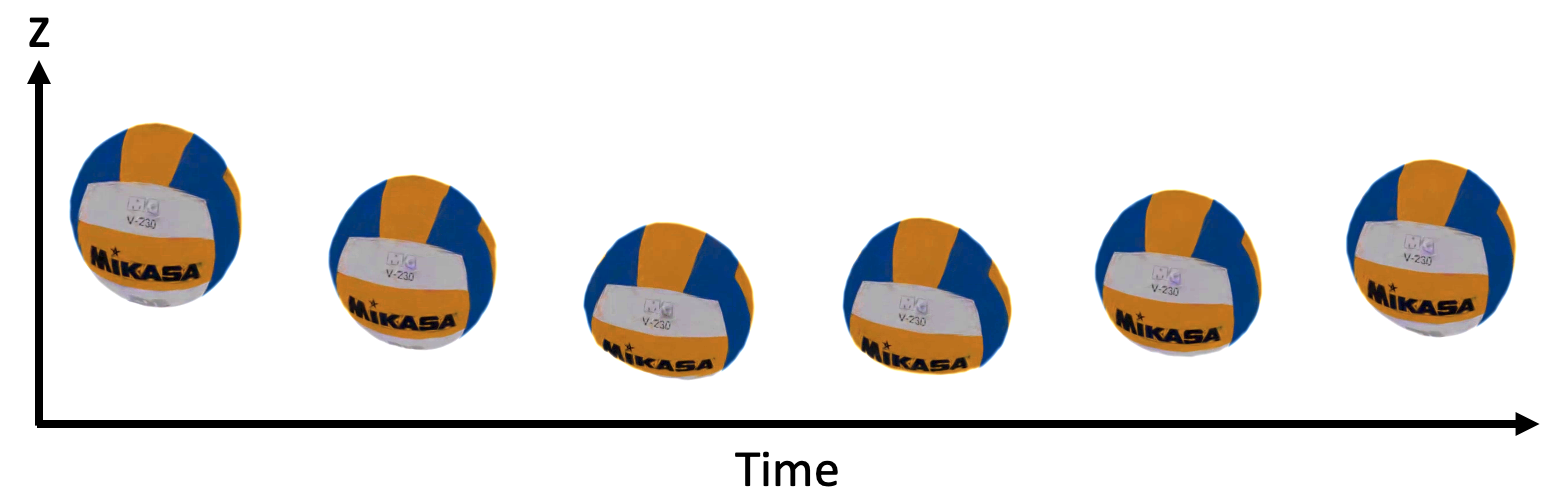}
    \caption{3D Gaussian Splatting integration with ODeform. The visualizations show our model's ability to transfer predicted deformations to Gaussian splats, enabling realistic rendering of physically-based animations.} 
    \label{fig:3dgs_backbone}
\end{figure}

In this experiment, we demonstrate our model's ability to transfer learned deformation dynamics to 3D gaussian splats. We begin with a textured volleyball mesh placed within a scanned 3D environment and render it from multiple camera viewpoints. Using these multi-view renderings, we reconstruct the scene with 3D Gaussian Splatting and isolate the region containing the volleyball.
We then apply our pre-trained mass-elastic deformation model (from Section \ref{exp_mass_bending}) to predict physically plausible deformations of the volleyball under various physical conditions. Since our model learns the deformation field of spherical objects given specific physical parameters, it can predict both local deformations and global position changes, which we use to reposition the Gaussian splats for subsequent rendering.
Figure \ref{fig:3dgs_backbone} illustrates the visual results of our approach, showing rendered splats deformed according to our model's predictions. This experiment highlights our model's versatility in generalizing to denser and noisier sparse representations beyond traditional meshes.


The combination of ODeform with 3DGS demonstrates the potential for integrating continuous deformation modeling to rendering techniques. Further exploration of this approach could be valuable for applications requiring both physical plausibility and visual fidelity. Additional qualitative results are provided in the supplementary material.

\subsection{Interpolation and Extrapolation}

\begin{table}[h!]
\centering
\caption{Extrapolation experiments on the Contact-Force and Mass-Elastic datasets. For the Contact-Force dataset, models are trained on Frames 0-10 and evaluated on 5 extrapolated frames. For the Mass-Elastic dataset, models are trained on Frames 0-20 and evaluated on the next 10 frames.}
\resizebox{\columnwidth}{!}{ 
    \begin{tabular}{lSS SS} \toprule
    \multirow{2}{*}{\textbf{Method}} & \multicolumn{2}{c}{\textbf{Contact-Force}} & \multicolumn{2}{c}{\textbf{Mass-Elastic}} \\ \cmidrule(lr){2-3} \cmidrule(lr){4-5}
     & {\textbf{RMSE} (mm) $\downarrow$} & {\textbf{MAE} (mm) $\downarrow$} & {\textbf{RMSE} (mm) $\downarrow$} & {\textbf{MAE} (mm) $\downarrow$} \\ \midrule
    nODE & 10.926 & 7.954 & 27.076 & 16.208 \\
    RNN & 27.302 & 20.107 & 28.718 & 18.169 \\
    \textbf{ODeform} & {\bfseries 9.261} & {\bfseries 7.051} & {\bfseries 3.597} & {\bfseries 2.440} \\ \bottomrule
    \end{tabular}
}
\label{tab:combined_extrapolation_results}
\end{table}

\begin{figure}[t]
    \includegraphics[width=\linewidth]{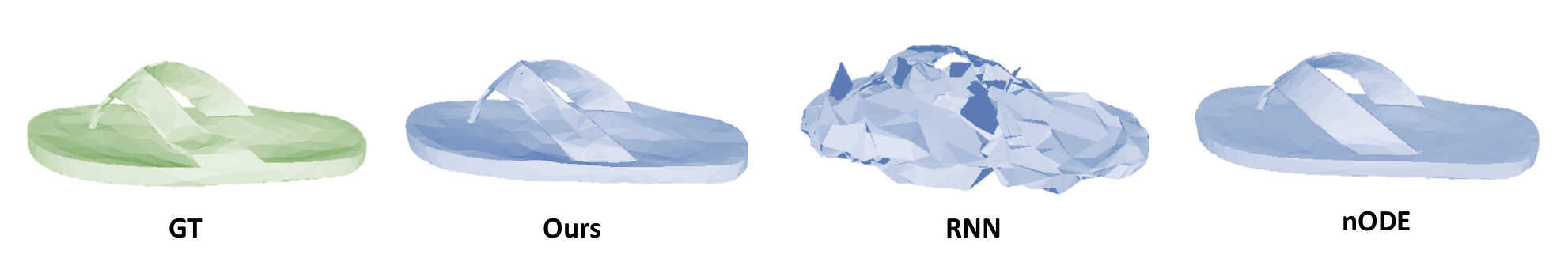}
    \caption{Extrapolation results on the FlipFlop object. Trained on first 10 frames our approach generates physically plausible deformations beyond the training timeframe (at t=15), while baseline methods fail to maintain coherent motion.} 
    \label{fig:extrapolation_flipflop}
\end{figure}

We present extrapolation performance across both datasets in Table \ref{tab:combined_extrapolation_results}, evaluating each method's ability to predict beyond the training sequence. For the Mass-Elastic dataset, models are trained on frames 0-20 and tested on 10 additional extrapolated frames. For the Contact-Force dataset, models are trained on frames 0-10 and evaluated on 5 extrapolated frames. Figure \ref{fig:extrapolation_flipflop} illustrates the extrapolation quality on the FlipFlop object with ODeform generating smooth, realistic deformations closer to the ground truth, while baseline models fail to predict the deformation and show distortions and artifacts. This setup assesses the methods' capacity to generalize temporally and maintain physically plausible deformations beyond their training distribution.

Additionally, to evaluate our model's ability to generalize and learn dynamics from sparse inputs at varying intervals, we designed an interpolation experiment using the Mass-Elastic Dataset. We trained our model using datasets where only every 3rd, 5th, or 8th frame was provided during training. %
Table~\ref{tab:interp_ball} presents the performance of our model under these conditions, against baselines. We compare against RNNs, nODE without latent encodings and OccuFlow \cite{niemeyer2019occupancy} pretrained network. 
For testing, the RNN, nODE and ODeform are provided with only the first frame of the test sequence along with its physical parameters. OccuFlow is given every 3rd, 5th, or 8th frame of the testing set, due to its design of interpolating between any two given frames. 
During the evaluation, the models predicted the complete test set without any skipped frames. 
Compared to the baselines our model consistently achieves the lowest RMSE, MSE, and MAE across all intervals (3, 5, and 8), demonstrating superior capacity to interpolate temporal data.

\begin{table}[ht]
\centering
\caption{Models trained on different intervals of frames on the Mass-Elastic dataset.}
\resizebox{\linewidth}{!}{%
\begin{tabular}{l c r r r} 
\toprule
\textbf{Method} & {\emph{\# Interval}} & \textbf{RMSE (mm)} $\downarrow$ & \textbf{MSE (mm)} $\downarrow$ & \textbf{MAE (mm)} $\downarrow$ \\ 
\midrule
\multirow{3}{*}{RNN}
    & 3 & 27.19 & 1010.00 & 17.42 \\
    & 5 & 28.37 & 1089.00 & 18.39 \\
    & 8 & 39.17 & 2056.00 & 25.06 \\ 
\midrule
\multirow{3}{*}{nODE}
    & 3 & 24.56 & 951.00  & 14.57 \\
    & 5 & 26.69 & 1133.00 & 15.89 \\
    & 8 & 26.59 & 1043.00 & 15.97 \\ 
\midrule
\multirow{3}{*}{OccuFlow\cite{niemeyer2019occupancy}}
    & 3 & 246.31 & 60899.00 & 207.43 \\
    & 5 & 244.42 & 59984.00 & 206.20 \\
    & 8 & 244.39 & 59937.00 & 206.60 \\ 
\midrule
\multirow{3}{*}{\textbf{Ours}}
    & 3 & \textbf{3.67} & \textbf{23.00}  & \textbf{2.56} \\
    & 5 & \textbf{5.05} & \textbf{38.00}  & \textbf{3.54} \\
    & 8 & \textbf{9.47} & \textbf{103.00} & \textbf{6.68} \\ 
\bottomrule
\end{tabular}
}
\label{tab:interp_ball}
\end{table}

\subsection{Parameter Optimization}
\label{sec:paramoptim}
We validate our model's understanding of physical dynamics through inverse parameter identification. Given a deformed observation, we leverage our trained model to reversely identify the physical parameters that caused it. By freezing the weights of our networks and optimizing only the physical parameters via gradient descent to minimize the difference between predicted and observed dynamics as follows:
\begin{equation}
\theta_{\text{new}} = \theta_{\text{old}} - \eta \nabla_{\theta} L(P(t)_{\text{pred}}, P(t)_{\text{gt}})
\end{equation}
we achieve low Mean Absolute Errors (MAE) of 0.113 for mass and 0.084 for bending, as shown in Table~\ref{tab:parameter_optim}. This demonstrates that our model not only predicts deformations from known parameters but also can effectively reverse-engineer physical parameters of a learned motion. The model has learned a meaningful latent representation of the underlying physics. This capability is valuable for real-world applications where material properties need to be estimated from observed deformations.

\begin{table}[ht]
\centering
\caption{Mean Absolute Error (MAE) for recovered physical parameters through reverse optimization.}
\resizebox{0.5\linewidth}{!}{
\begin{tabular}{lSS} \toprule
    \textbf{Metric} & \textbf{Mass} & \textbf{Bending} \\ \midrule
    \textbf{MAE} & 0.113 & 0.084 \\ \bottomrule
\end{tabular}}
\label{tab:parameter_optim}
\end{table}





\section{Conclusion}
Despite recent advances in 3D shape representation, accurately modeling physically consistent, continuous deformations remains a significant challenge. Our approach addresses this by leveraging neural ODEs to represent deformation as a continuous flow within a learned embedding space, enabling accurate prediction across arbitrary timepoints from sparse observations. This continuous representation, rather than discrete states, enables smoother interpolation, more accurate extrapolation, and better generalization to unseen physical configurations. Beyond improved accuracy, our framework enables novel capabilities like parameter optimization and generalization across 3D representations. Looking forward, extending the method to multi-object interactions presents exciting opportunities for applications in robotic manipulation and simulation.

\bibliographystyle{IEEEtran}

{
\bibliography{main}
}

\addtolength{\textheight}{-12cm}   
\end{document}